\documentclass[conference]{IEEEtran}
\IEEEoverridecommandlockouts
\usepackage{cite}
\usepackage{amsmath,amssymb,amsfonts}
\usepackage{algorithmic}
\usepackage{graphicx}
\usepackage{textcomp}
\usepackage{xcolor}
\usepackage{numprint} 
\usepackage{caption}
\usepackage{multirow}
\usepackage{float}
\usepackage[table]{xcolor}
\usepackage{caption}  
\usepackage[bookmarksnumbered, colorlinks, plainpages, linkcolor=black, citecolor=black, urlcolor=black]{hyperref}
\usepackage{tikz,xcolor,hyperref}
\definecolor{lime}{HTML}{A6CE39}
\DeclareRobustCommand{\orcidicon}{
	\begin{tikzpicture}
	\draw[lime, fill=lime] (0,0) 
	circle [radius=0.16] 
	node[white] {{\fontfamily{qag}\selectfont \tiny ID}};
	\draw[white, fill=white] (-0.0625,0.095) 
	circle [radius=0.007];
	\end{tikzpicture}
	\hspace{-2mm}
}
\foreach \x in {A, ..., Z}{\expandafter\xdef\csname orcid\x\endcsname{\noexpand\href{https://orcid.org/\csname orcidauthor\x\endcsname}
			{\noexpand\orcidicon}}
}

\usepackage[letterpaper, top=72pt, bottom=54pt, left=54pt, right=54pt]{geometry}
\def\BibTeX{{\rm B\kern-.05em{\sc i\kern-.025em b}\kern-.08em
    T\kern-.1667em\lower.7ex\hbox{E}\kern-.125emX}}
\begin{document}

\title{LiDAR Point Cloud Image-based Generation Using Denoising Diffusion Probabilistic Models}

\newcommand{\orcidauthorB}{0009-0007-2690-4496} 
\newcommand{\orcidauthorE}{0000-0002-5211-3598}
\author{Amirhesam Aghanouri\orcidB \emph{Member, IEEE}, Cristina Olaverri-Monreal\orcidE{} \emph{Senior Member, IEEE}%
\thanks{Johannes Kepler University Linz, Austria; Department Intelligent
 Transport Systems
\texttt{\{amirhesam.aghanouri, cristina.olaverri-monreal\}@jku.at}}}

\maketitle

\begin{abstract}
Autonomous vehicles (AVs) are expected to revolutionize transportation by improving efficiency and safety. Their success relies on 3D vision systems that effectively sense the environment and detect traffic agents. Among sensors AVs use to create a comprehensive view of surroundings, LiDAR provides high-resolution depth data enabling accurate object detection, safe navigation, and collision avoidance. However, collecting real-world LiDAR data is time-consuming and often affected by noise and sparsity due to adverse weather or sensor limitations. This work applies a denoising diffusion probabilistic model (DDPM), enhanced with novel noise scheduling and time-step embedding techniques to generate high-quality synthetic data for augmentation, thereby improving performance across a range of computer vision tasks, particularly in AV perception. These modifications impact the denoising process and the model's temporal awareness, allowing it to produce more realistic point clouds based on the projection. The proposed method was extensively evaluated under various configurations using the IAMCV and KITTI-360 datasets, with four performance metrics compared against state-of-the-art (SOTA) methods. The results demonstrate the model's superior performance over most existing baselines and its effectiveness in mitigating the effects of noisy and sparse LiDAR data, producing diverse point clouds with rich spatial relationships and structural detail.

\end{abstract}

\begin{IEEEkeywords}
Diffusion models, Generative modeling, LiDAR scene generation, AV perception, Data augmentation.
\end{IEEEkeywords}  

\section{Introduction}
\label{sec:Introduction}
The reliability of autonomous vehicles (AVs) must be demonstrated across complex and dynamic scenarios, such as urban intersections, highway merges, varying weather, and congested streets \cite{bib:olaverri2020promoting}. Successfully navigating these situations relies on strong 3D perception systems that integrate data from multiple sensors, including cameras, radio detection and ranging (RADAR), and light detection and ranging (LiDAR). Compared to other sensors, LiDAR can be used in various weather and lighting conditions. It measures depth by emitting laser pulses and measuring their return time, resulting in a comprehensive point cloud data mapping three-dimensional surroundings. This high-resolution spatial data helps AVs detect and classify objects from a distance, ensuring timely responses to hazards. Point cloud data captured by LiDAR sensors is often incomplete and noisy due to various factors, including the shape and texture of objects, adverse weather conditions like rain or snow, issues with surface reflectivity, and inherent limitations of the sensors. These factors distort important geometric information that can reduce the performance of subsequent analyses and applications. To address issues during data recording, several approaches can be considered. The first involves increasing the number of LiDAR sensors that is often impractical due to high costs and power consumption. Another approach is sensor fusion which enhances point cloud quality by integrating data from multiple sensors. However, it can be complex and costly, requiring careful calibration and synchronization to ensure perfect data fusion and prevent issues like misalignment or adding shadows to agents.

An efficient alternative is to use deep generative models to generate high-quality and diverse LiDAR point clouds without relying on additional hardware. These synthetic datasets help create more complete and noise-free training data which is particularly valuable when real-world LiDAR data is scarce, expensive, or challenging to collect, improving the accuracy and robustness of perception models. This paper proposes a method for generating point clouds through bird’s-eye view (BEV) and equirectangular projections. While BEV helps in spatial understanding, the equirectangular format retains full depth and angular information, enabling precise reconstruction of 3D environments. To this end, denoising diffusion probabilistic models (DDPMs) \cite{bib:ho2020denoising} were selected due to their strong performance in data restoration tasks. Unlike other generative models such as variational autoencoders (VAEs) \cite{bib:kingma2013auto} which may suffer from posterior collapse or generative adversarial networks (GANs) \cite{bib:goodfellow2014generative} which are prone to mode collapse and unstable training, DDPMs iteratively refine noisy data through a learned denoising process. This makes them well-suited for reconstructing detailed geometric structures critical for 3D perception. In this work, DDPMs were further adapted and enhanced using novel techniques to improve the quality of the generated results for autonomous driving scenarios that require large and diverse datasets. For reliable training and evaluation, two benchmark datasets were used, the interaction of autonomous and manually controlled vehicles (IAMCV) \cite{bib:certad2024iamcv, bib:d1g3-c160-23} and the KITTI-360 dataset \cite{bib:liao2022kitti}. Multiple configurations were tested using four evaluation metrics and the most effective setup was identified and compared with SOTA methods to evaluate its overall performance and real-world applicability.

The following section begin with an overview of recent deep generative models and their limitations. Section \ref{sec:Methodology} introduces the proposed model, key improvements, and the datasets used for evaluation. Section \ref{sec:Approach evaluation} describes the evaluation information, implementation details, and baseline comparisons. The experimental results are presented in Section \ref{sec:Results}, followed by conclusions and future work in Section \ref{sec:Conclusion and Future work}.

\section{Related literature}
\label{sec:Related literature}
LiDAR simulation methods are important for testing AVs and robotic algorithms, reducing risks in real-world trials. The car learning to act (\textit{CARLA}) platform \cite{bib:dosovitskiy2017carla} is widely used for simulating sensor setups and scenarios to evaluate AV driving strategies. However, \textit{CARLA} and similar tools face challenges, including difficulty in generating 3D models, reliance on real-world scans, and high computational demands of physics-based simulations. To address these, deep generative models are explored for synthetic LiDAR data generation to train AV perception systems. Earlier methods like LiDAR GAN and LiDAR VAE \cite{bib:caccia2019deep} projected point clouds into 2D range images. Techniques such as interpolation and noise addition mitigated issues like LiDAR ray drop caused by missing data due to reflections but challenges such as fuzzy or missing details were often exhibited in the generated results due to inherent limitations in the frameworks. UltraLiDAR \cite{bib:xiong2023learning} is an another framework for BEV image generation and manipulation, utilized the vector-quantized VAE (VQ-VAE) model to create a compact and discrete representation of 3D LiDAR data. However, due to the large size and sparsity of point clouds in this view, the method spends significant computational resources on generating empty voxels which slowed down generation speed. LiDARGen \cite{bib:zyrianov2022learning} was proposed as a further approach that generated point clouds through a progressive denoising process based on a score-based method. While it provided physically accurate data, it encountered limitations due to its sampling-based strategy. LiDAR diffusion models (LiDMs) \cite{bib:ran2024towards} were also introduced to address challenges in preserving complex structural details and object geometries. Multimodal conditioning enabled the model to rely on diverse inputs, enhancing its applicability, although the requirement of eight 24GB GPUs for training posed limitations in terms of accessibility and reproducibility. Unlike previous work, this study introduces entirely new approaches to noise scheduling and time-step embedding in the generation process of DM. Furthermore, the model was evaluated using a new benchmarking dataset that supports both 2D and 3D outputs. These innovations enhance output detail, simplify model design, generate consistent point clouds, and offer a user-friendly deployment framework.

\section{Methodology}
\label{sec:Methodology}
\begin{figure*}
    \centering
    \includegraphics[width=0.9\textwidth]{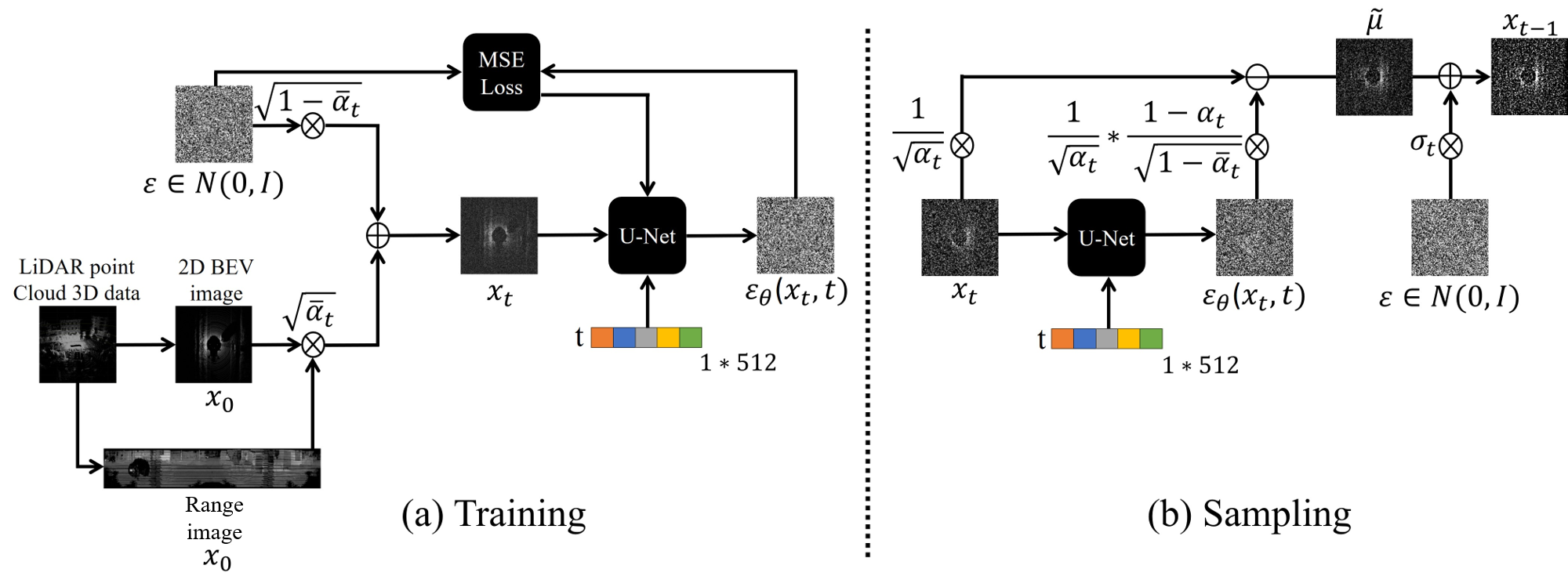}
    \caption[Schematic representation of the training and sampling process of the DDPM.]{Schematic representation of the training and sampling process of the DDPM.}
    \label{fig:training_sampling}
\end{figure*}

The method presented in this paper is DDPM, a diffusion model operating in discrete time for training and sampling \cite{bib:sohl2015deep}. Unlike GANs which generate data in one step, DDPM uses multiple steps to capture information more effectively, resulting in higher-quality outputs. As part of the training process, during the forward diffusion process, Gaussian noise ($\varepsilon \sim \mathcal{N}(0, \mathbf{I})$) is added to the image distribution $\mathbf{x}_0$ (BEV or equirectangular) \cite{bib:ho2020denoising} to create a noisy image ($\mathbf{x}_t$), gradually degrading the signal. Noise addition is controlled by a scheduler $\beta_t = 1 - \alpha_t$ and $\bar{\alpha}_t=\prod_{i=1}^{t} \alpha_i$ is the signal strength of the image in each step. Using U-Net \cite{bib:ronneberger2015u} as the neural network ($p_{\theta}$), the added noise at each step is predicted as $\epsilon_\theta(\mathbf{x}_t, t)$. The mean squared error (MSE) between the forward process noise and the model-predicted noise is calculated and the model's parameters $\mu_\theta$ and $\Sigma_\theta$ are updated during the training, as shown in part (a) of \autoref{fig:training_sampling}. As part of the sampling process, once the model has learned to predict noise accurately, new samples can be generated. The process begins with random noise drawn from a standard normal distribution. The model then iteratively denoises the data, moving backward from the final step $T$ to step 1. At each step, predicted noise is subtracted and new noise is added to prevent mode collapse by maintaining stochasticity and ensuring sample diversity, as shown in part (b) of \autoref{fig:training_sampling}, except between steps 1 and 0, where no additional noise is introduced due to a noise-free output.

\subsection{Noise Schedule Design}
\label{sec:Methodology:Noise Schedule Design}
The performance of DDPM is influenced by noise scheduling which controls the noise intensity at each step. Imperfections in the reverse process occur when steps with a low signal-to-noise ratio (SNR) fail to improve sample quality. A schedule designed for smaller resolutions may not be effective for higher resolutions, as larger images with pixel redundancy are less sensitive to information loss at similar noise levels. While scaling the image can benefit some datasets, it negatively impacts the quality of generated samples here due to the sensitivity of point clouds. To address this, all introduced noise scheduling functions were tested to determine which one offers better accuracy. Among the constant, linear, and quadratic schedules introduced in baseline DDPMs \cite{bib:ho2020denoising}, the constant and linear schedules produced low-quality samples due to rapid SNR degradation. With 1000 diffusion steps, the SNR approached zero after the 600th step, causing these steps to be skipped and reducing the model's ability to capture additional details. The quadratic schedule mitigated signal degradation and preserved more details during the diffusion process but the results still did not achieve the desired quality. The \(\text{cosine}^2\) schedule \cite{bib:nichol2021improved} which features a linear SNR drop in the middle of the process and maintains stability at the tails to address rapid signal degradation, failed to produce sufficient sample quality due to the higher resolution variations tested. The sigmoid schedule \cite{bib:jabri2022scalable} struggled to produce clean datasets due to the complexity of its hyperparameters. Hyperbolic schedule \cite{bib:sohl2015deep} which linearly diminishing signal strength, outperformed the aforementioned approaches but still did not achieve the desired level of performance required to compete with SOTA methods, even after different hyperparameter tuning.

\begin{figure}[htbp]
    \centering
    \includegraphics[width=0.48\textwidth]{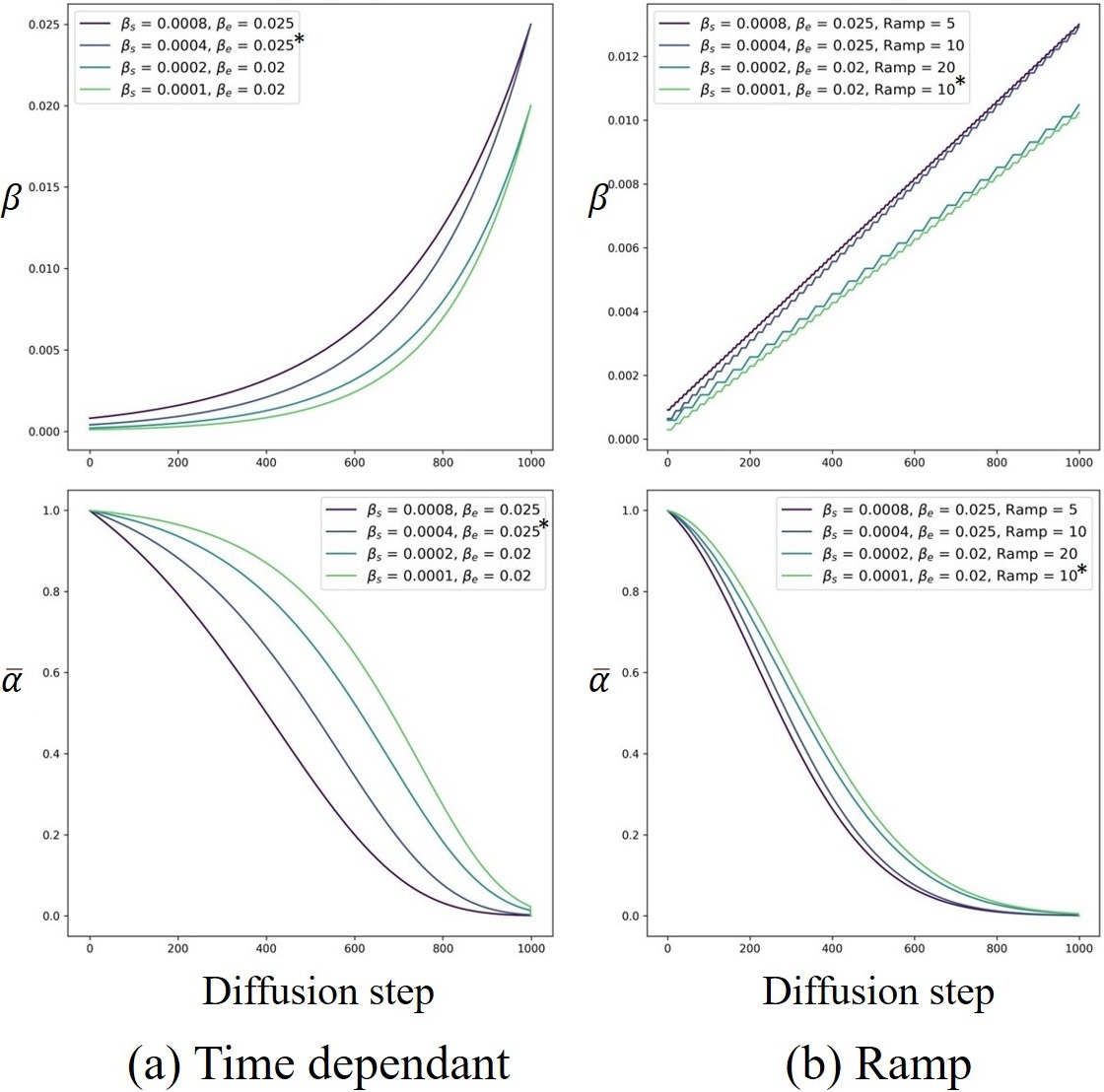}
    \caption[Impact of time-dependent and ramp schedules on noise and signal.]{Impact of time-dependent and ramp schedules on noise and signal. The upper plots show noise variance changes with increasing \(\beta\) and optimal hyperparameters marked by stars, while the bottom plots show signal strength decreasing over time.}
    \label{fig:new_noise_schedules}
\end{figure}

\begin{figure}[htbp]
    \centering
    \includegraphics[scale=0.42]{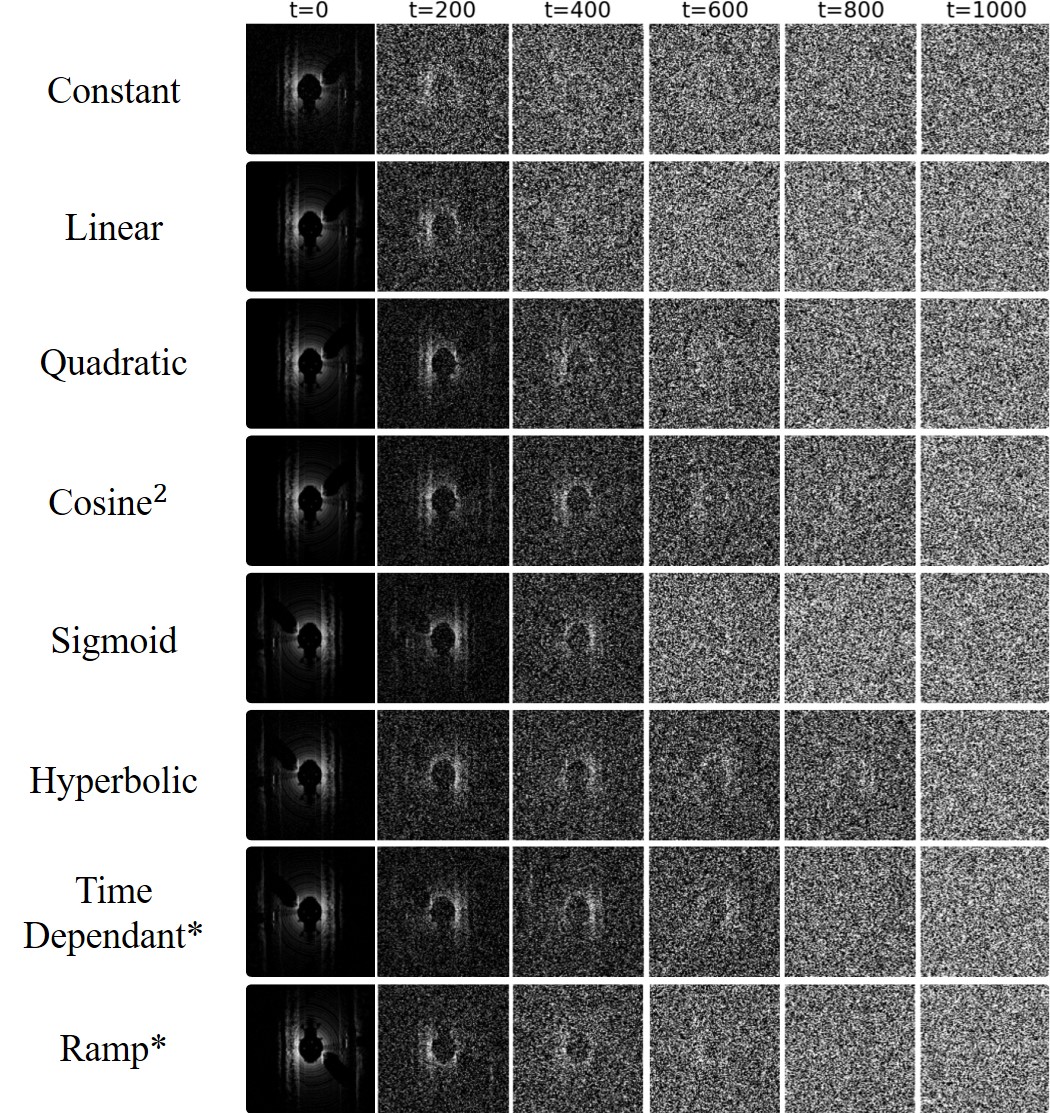}
    \caption[Effect of different noise schedules during the forward process on image SNR]{Effect of different noise schedules during the forward process on image SNR. Signal degradation should occur gradually, like in a time-dependant method.}
    \label{fig:all_noise_schedules}
\end{figure}

In this study, two new noise schedules, time-dependant and ramp were implemented for the first time to increase the model's performance. Their impacts on noise and signal in the forward process are visualized in \autoref{fig:new_noise_schedules}. Furthermore, \autoref{fig:all_noise_schedules} illustrates the effects of all schedules on SNR over 1000 steps, highlighting how signal reduces rapidly. The new schedules were shown to outperform existing ones on our datasets by effectively managing noise removal across scales and the ability to reduce the number of sampling time-steps from 1000 to 800. They keep essential image information from degrading rapidly over time while ensuring near-zero SNR in the final forward step.

Time-dependant schedule characterized by an exponential graph with an increasing slope for $\beta_{t}$. It helped to reduce the number of sampling steps without significantly compromising the quality of the generated samples because the model was trained on a more stable noise schedule and its progression was more manageable. This schedule proved its effectiveness

across various image resolutions used in this work. $\beta_{t}$ is computed using the following formula:

\begin{equation}
\beta_t = \beta_{start} * (\frac{\beta_{end}}{\beta_{start}})^{timing}
\end{equation}

where $timing \in [0,1]$ increases linearly over time-steps. The second noise schedule introduced here is the ramp schedule which combines constant and linear noise, showing increased robustness and producing high-quality samples. The constant segment defined by the first step's value in the linear section, extends to the last time-step. The trajectory of $\beta$ for this noise schedule is calculated:

\begin{equation}
\beta_t = \beta_{\text{start}} + \frac{\beta_{\text{end}} - \beta_{\text{start}}}{T-1} \cdot t, \quad t \in [0, T-1]
\end{equation}
\begin{equation}
\begin{aligned}
\beta_{0:T} = \bigcup_{k=0}^{L-1} \Big( 
&\underbrace{\beta_{10k+10}, \beta_{10k+10}, \dots, \beta_{10k+10}}_{\text{constant part}}, \\
&\underbrace{\beta_{10k+10}, \beta_{10k+11}, \dots, \beta_{10k+19}}_{\text{linear part}} 
\Big)
\end{aligned}
\end{equation}

where $T$ represents the total number of time-steps, \( L = \frac{T}{2n} \) is the number of ramp shapes, and \( n \) is the segment length.

\subsection{Time-step Embedding}
\label{sec:Methodology:Time-step Embedding}
In the reverse process, the U-Net model takes two inputs: an image and an encoded time-step which informs the step and noise level to the model, enabling precise predictions. Each step is represented by a continuous vector to prevent numerical instability caused by large raw integer values in the U-Net model. Converting time-steps ensures smooth generalization and easier interpolation between nearby steps. It also helps the model capture fine and coarse temporal patterns, enabling recognition of small changes between steps and larger trends over time. In this work, a novel approach introduced for the first time using Fourier series to improve the model's understanding of time-steps for more accurate noise prediction, alongside testing sinusoidal positional embeddings introduced in \cite{bib:vaswani2017attention}. The Fourier series offers a more expressive and flexible representation of time-steps which is crucial for capturing noise progression and denoising across time-steps. By adjusting the number of harmonics, the capture of time-step information is fine-tuned, enabling DDPM to handle variations in denoising rates better than a simple sinusoidal function. Unlike sinusoidal functions which impose strict periodicity, Fourier series accommodates noise dynamics by combining periodic components. The general formula for this new embedding is as follows:

\begin{equation}
\text{TE}(t, i) = \sum_{n=1}^{N} \frac{1}{n} \left( \sin\left(\frac{n \cdot t}{10000^{\frac{2i}{d}}}\right) + \cos\left(\frac{n \cdot t}{10000^{\frac{2i}{d}}}\right) \right)
\label{eq:fourier}
\end{equation}

where time-step \( t \) is expressed as integers, while \( i \) is the dimension index, ranging from 0 to \( d/2 \). The parameter \( d \) is total embedding dimension and typically even number. By alternating sine and cosine functions, the embeddings gain orthogonal properties, enhancing the uniqueness of positional vectors. The number of harmonics \( N \) captures fine-grained and coarse-grained details for each vector element corresponding to the time-step. The level of detail is controllable by adjusting the number of harmonics $N$. The term \( 10000^{\frac{2i}{d}} \) determines the sinusoidal frequencies. As \( i \) increases, the divisor grows, reducing frequency and extending wavelength. The wavelengths form a geometric progression from \( 2\pi \) to \( 20000\pi \), with 10,000 as the base. Finally, each time-step embedding is transformed into higher dimensions using learnable linear layers, optimizing parameters during training.

\subsection{Datasets}
\label{sec:Methodology:Datasets}
The model was trained and evaluated on two datasets: IAMCV and KITTI-360. The IAMCV dataset, focused on vehicle interactions was used for the first time in generative modeling. A subset comprising \(\sim77,300\) LiDAR frames collected in Germany with three LiDAR sensors was used. The central sensor captured objects up to 240 meters under various conditions. 80\% of the dataset was used for training, 10\% for validation, and 10\% for testing, ensuring proper distribution for evaluation. The KITTI-360 dataset includes nine sequences of suburban scenes from Karlsruhe, recorded using sensors capturing 360-degree views. A total of 76,714 frames were used, with the same distribution as the IAMCV dataset for training, validation, and testing.

To enable structured learning from LiDAR data, raw 3D point clouds were projected into two 2D representations, BEV and equirectangular view. The BEV was generated through an orthographic projection along the Z-axis, providing a top-down view with a resolution of 1024×1024. In this view, pixel values were determined by the intensity values of the projected points. For the equirectangular projection, following \cite{bib:zyrianov2022learning}, each 3D point \((x, y, z)\) was mapped to spherical coordinates \((\theta, \phi, d)\), where \(d = \sqrt{x^2 + y^2 + z^2}\), \(\theta = \arccos(z / \sqrt{x^2 + y^2 + z^2})\), and \(\phi = \text{atan2}(y, x)\). \(\theta\) denoted the inclination, \(\phi\) the azimuth, and \(d\) the depth values that were normalized to the range [0, 1]. The resulting equirectangular image was represented as a 64×1024 grid. As a contribution, a local smoothing operation was applied to the depth values prior to encoding them into the equirectangular format. This operation reduced some noise in sparse regions by applying a Gaussian filter over local neighborhoods of points. The filter size was adapted based on local point density which was estimated using a k-nearest neighbors algorithm with optimized \(k = 10\), preserving fine details in areas of high point density. The equirectangular images were also back-projected into 3D space using the inverse spherical-to-cartesian transformation, enabling direct comparison with the original LiDAR point clouds.

\section{Approach evaluation}
\label{sec:Approach evaluation}
\textit{1) Metrics:} The performance of the generative model was evaluated using four key metrics, as described in \cite{bib:ran2024towards}. For statistical evaluation, Jensen-Shannon divergence (JSD) \cite{bib:lin1991divergence} and maximum mean discrepancy (MMD) \cite{bib:gretton2007kernel} were used to assess the quality of BEV representations. JSD was used to quantify the similarity between the pixel intensity histograms of real and generated images, providing a symmetric and bounded measure of divergence between two probability distributions. MMD, a kernel-based method was applied to compare distributions using the radial basis function (RBF) kernel, capturing subtle differences that may affect navigation consistency. For perceptual evaluation, Fréchet range image distance (FRID) and Fréchet point-based volume distance (FPVD) were used \cite{bib:ran2024towards}. FRID was used to measure the semantic consistency of range images by comparing feature representations extracted from pre-trained segmentation networks which reduced sampling variability and focused on high-level semantic features. FPVD was used to evaluate the geometric fidelity of 3D point clouds by comparing features extracted from point-based networks, capturing differences in the geometric structure of the generated point clouds.

\textit{2) Implementation details:} DDPM with a U-Net backbone was trained to generate firstly images at both $1024 \times 1024$ and $64 \times 1024$ resolutions. A dropout rate of 0.1 was used to enhance regularization. Training was conducted using the AdamW optimizer with a learning rate of $2 \times 10^{-4}$ combined with MSE loss. A batch size of 4 was used and although the model was initially trained for 50 epochs, early stopping was applied depending on the configuration. Each epoch took between 12 to 16 minutes on a setup of one NVIDIA RTX A6000 GPU. At inference, each sample was generated in 1.68 to 2.06 seconds with the model comprising 16,626,049 trainable parameters.

\textit{3) Downstream evaluation:} To evaluate the plausibility of the generated 3D point clouds, a pretrained RangeNet++ model \cite{milioto2019rangenet++} was used for semantic segmentation and trained on the SemanticKITTI dataset \cite{behley2019semantickitti}. This approach allowed for the assessment of how well the pretrained model could predict semantic labels for the synthetic point clouds, verifying whether the generated data contained realistic semantic structures similar to those found in real-world LiDAR scans.

\textit{4) Baselines:} Several baselines were considered for comparison, included VAE-based models, such as LiDAR VAE and UltraLiDAR, as well as the GAN-based LiDAR GAN. In addition, diffusion-based methods, including LiDARGen and LiDMs were incorporated to ensure a comprehensive evaluation across diverse generative frameworks.

\section{Results}
\label{sec:Results}
Sixty-four potential configurations were tested, derived from combinations of two time-step embedding techniques, two projection methods, two datasets, and eight noise schedules. Applied to the DDPM model, 11 configurations were chosen for presentation according to performance, with details provided in \autoref{tab:all_configurations}. A quantitative comparison was made between our model and the SOTA models on KITTI-360 and the optimal configuration for IAMCV which is considered for the first time here is also identified. The score metrics for the selected configurations and SOTA methods on the both datasets based on 800 and 1000 randomly generated samples are shown in \autoref{tab:quantitative}.

\begin{figure}[H]
    \centering
    \includegraphics[scale=0.42]{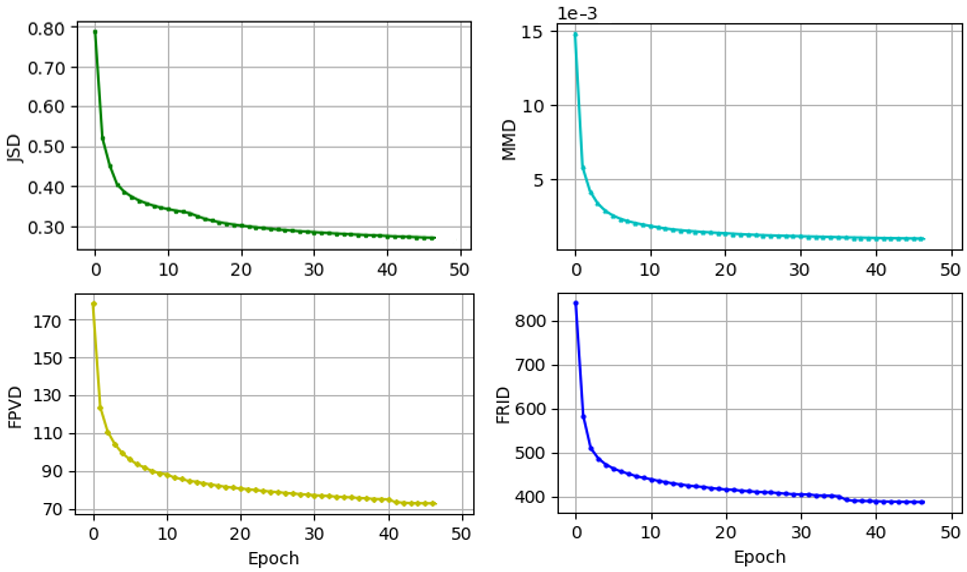}
    \caption[Convergence metrics across training epochs for configuration "J".]{Convergence metrics across training epochs for configuration "J".}
    \label{fig:learning_curves}
\end{figure}

Configuration "K" shown in brown, achieved the best performance on the IAMCV dataset, validating the effectiveness of time-dependent noise scheduling and the incorporation of Fourier series-based time-step embeddings. Configuration "J" shown in blue has the same settings as "K" and demonstrated SOTA performance on the KITTI-360 dataset, surpassing all existing baselines except LiDMs \cite{bib:ran2024towards} which is shown in green. \autoref{fig:learning_curves} shows the score metrics for configuration "J" during training which stopped at epoch 47. These settings performed well at accuracy and stability. Although LiDMs reports higher accuracy because it benefits from a larger backbone that made it feasible by training on significantly more powerful GPU hardware.

\begin{table}[h]
\centering
\caption{Characteristics of the selected configurations}
\begin{tabular}{|>{\centering\arraybackslash}p{1.6cm}|>{\centering\arraybackslash}p{2cm}|>{\centering\arraybackslash}p{1.6cm}|>{\centering\arraybackslash}p{1.3cm}|}
\hline
\textbf{Configuration} & \textbf{Schedule} & \textbf{Embedding} & \textbf{Dataset}\\
\hline
\textbf{A} & Linear & Sinusoidal & IAMCV \\
\textbf{B} & Linear & Fourier series & KITTI-360 \\
\textbf{C} & Hyperbolic & Fourier series & IAMCV \\
\textbf{D} & Hyperbolic & Sinusoidal & KITTI-360 \\
\textbf{E}* & Ramp & Fourier series & KITTI-360 \\
\textbf{F} & Ramp & Fourier series & KITTI-360 \\
\textbf{G} & Ramp & Sinusoidal & IAMCV \\
\textbf{H} & Time-dependant & Sinusoidal & IAMCV \\
\textbf{I} & Time-dependant & Sinusoidal & KITTI-360 \\
\textbf{J} & Time-dependant & Fourier series & KITTI-360 \\
\textbf{K} & Time-dependant & Fourier series & IAMCV \\
\hline
\end{tabular}
\label{tab:all_configurations}
\end{table}

\begin{table}[h!]
\centering
\caption{Quantitative results on both datasets}
\begin{tabular}{|>{\centering\arraybackslash}p{2.06cm}|>{\centering\arraybackslash}p{0.75cm}|>{\centering\arraybackslash}p{1.35cm}|>
{\centering\arraybackslash}p{0.93cm}|>
{\centering\arraybackslash}p{0.99cm}|}
\hline
\textbf{Model} & \textbf{JSD $\downarrow$} & \textbf{MMD $\downarrow$} & \textbf{FRID $\downarrow$} & \textbf{FPVD $\downarrow$}\\
\hline
\rowcolor{red!70}\textbf{Full noise} & 0.472 & $429 \times 10^{-5}$ & 3614 & 374 \\
\textbf{A} & 0.349 & $316 \times 10^{-5}$ & 1471 & 257 \\
\textbf{B} & 0.338 & $303 \times 10^{-5}$ & 1365 & 242 \\
\textbf{C} & 0.255 & $244 \times 10^{-5}$ & 735 & 179 \\
\textbf{D} & 0.251 & $239 \times 10^{-5}$ & 688 & 171 \\
\rowcolor{yellow!70}\textbf{E*} & 0.224 & $208 \times 10^{-5}$ & 502 & 124 \\
\textbf{F} & 0.215 & $192 \times 10^{-5}$ & 436 & 109 \\
\textbf{G} & 0.227 & $214 \times 10^{-5}$ & 527 & 133 \\
\textbf{H} & 0.213 & $190 \times 10^{-5}$ & 421 & 101 \\
\textbf{I} & 0.211 & $185 \times 10^{-5}$ & 413 & 96 \\
\rowcolor{blue!40}\textbf{J} & 0.201 & $161 \times 10^{-5}$ & 359 & 78 \\
\rowcolor{brown!70}\textbf{K} & 0.208 & $176 \times 10^{-5}$ & 376 & 92 \\
\textbf{LiDAR GAN} \cite{bib:caccia2019deep} & 0.284 & $266 \times 10^{-5}$ & 1143 & 197 \\
\textbf{LiDAR VAE} \cite{bib:caccia2019deep}& 0.247 & $227 \times 10^{-5}$ & 702 & 153 \\
\textbf{LiDARGen} \cite{bib:zyrianov2022learning}& 0.229 & $215 \times 10^{-5}$ & 593 & 145 \\
\textbf{UltraLiDAR} \cite{bib:xiong2023learning}& 0.213 & $188 \times 10^{-5}$ & 426 & 100 \\
\rowcolor{green!70}\textbf{LiDMs} \cite{bib:ran2024towards}& 0.198 & $148 \times 10^{-5}$ & 324 & 69 \\
\hline
\end{tabular}
\label{tab:quantitative}
\end{table}

\begin{figure}[H]
    \centering
    \includegraphics[scale=0.24]{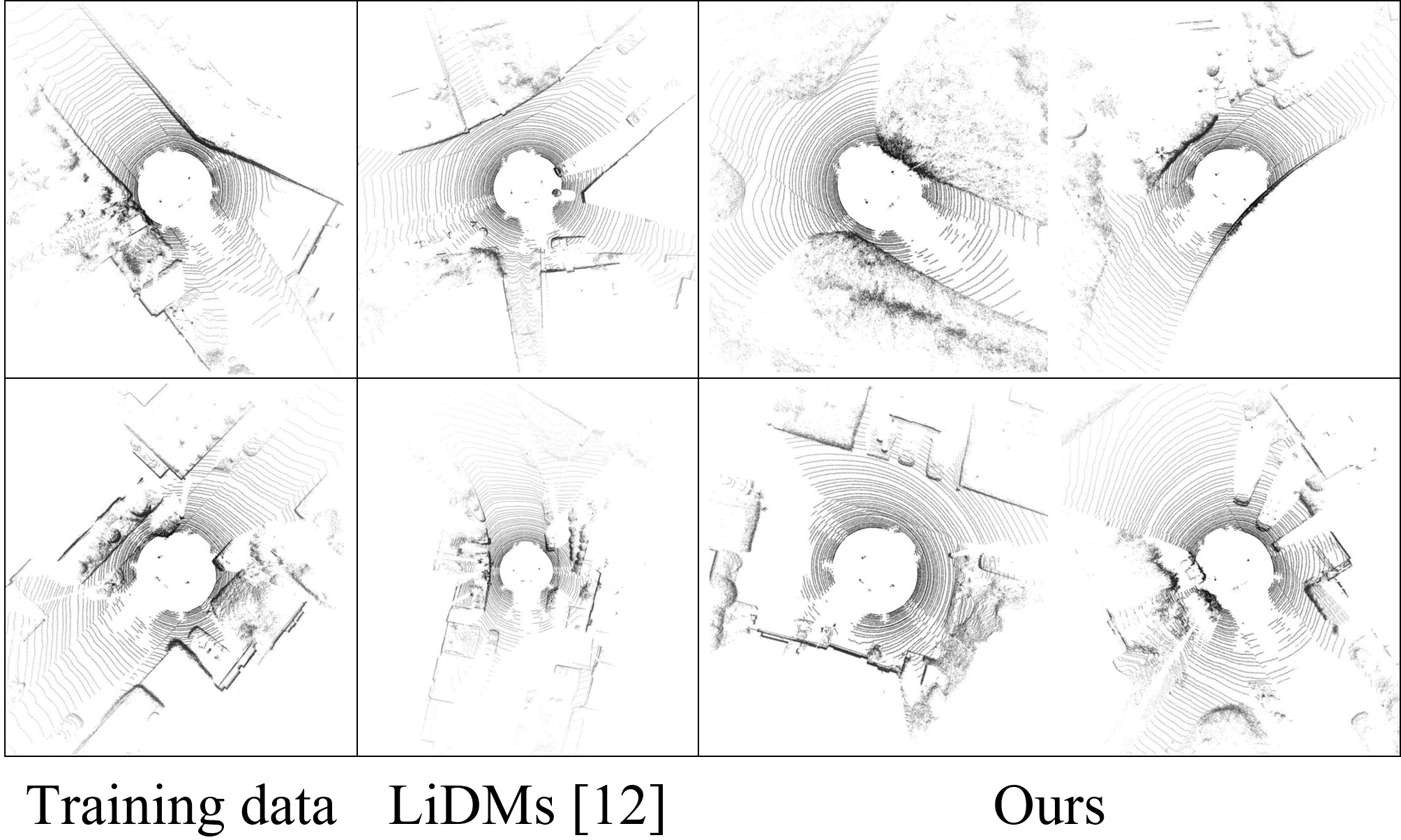}
    \caption[Comparison of LiDAR point cloud generation methods.]{Comparison of LiDAR point cloud generation methods on KITTI-360 dataset.}
    \label{fig:samples}
\end{figure}

In contrast, our approach is intentionally designed around a lightweight U-Net backbone, prioritizing architectural simplicity for easier implementation and lower computational cost. While the performance gap can be attributed to methodological limitations, the difference in model capacity also plays a significant role. Configuration "E" marked with a star in \autoref{tab:all_configurations} and shown in yellow in the table above, shares the same settings as "F" but uses 800 diffusion steps for sampling. This highlights the effectiveness of the new noise schedule which also enables faster sampling approximately 250 to 300 milliseconds faster per sample compared to the 1000 steps used in the other configurations which typically require between 1.68 and 2.06 seconds per sample at $64 \times 1024$ resolution with our GPU. Although there is still space for optimization, since LiDMs takes about 1.35 seconds per sample and is faster. Results from other configurations, such as "A", "B", "C", and "D" highlight the rationale behind introducing the new noise schedule and time-step embedding methods to enhance our model's performance and competitiveness with SOTA methods. The results from the semantic segmentation indicated that meaningful labels were assigned by the model. The mean intersection over union (mIoU) for the generated point clouds was found to be 47.4\% which is in close proximity to the 52.2\% mIoU achieved by RangeNet++ on the real-world data from the test set. Despite the slight mIoU gap, the synthetic point clouds maintain realistic semantic distributions and can be used effectively for downstream tasks such as training or simulation. In \autoref{fig:samples}, generated samples from the KITTI-360 dataset are shown. Real data, the best baseline, and the proposed method are compared. High-fidelity LiDAR point cloud structures are achieved by the proposed method, both locally and globally with results that closely resemble the ground truth.

\section{Conclusion and Future work}
\label{sec:Conclusion and Future work}
This study explored the potential of DDPMs to generate detailed and accurate point cloud data. The proposed methodology demonstrated the model's ability to handle the complexity and sensitivity of LiDAR data, producing realistic outputs with diverse characteristics. The application of novel noise schedules and Fourier-based time-step embeddings significantly contributed to the model's accuracy and efficiency, surpassing SOTA methods. The generated results exhibited high realism, making them ideal for tasks such as image segmentation and object detection that require large datasets to enhance the perception systems of AVs. Future work should focus on exploring alternative noise distributions and integrating other diffusion model variants to further optimize performance. Techniques such as learnable noise schedules and super-resolution can help address computational and quality-related challenges.

\section{Acknowledgement}
\label{sec:Acknowledgement}
This work was partially supported by the Austrian Research Promotion Agency (FFG) pDrive OOE 2022 - Future Mobility - Partnerantrag, project number: 901692
\bibliographystyle{IEEEtran}
\bibliography{bliblio}

\end{document}